\documentclass[11pt]{article}

\usepackage[preprint]{acl}

\usepackage{times}
\usepackage{latexsym}
\usepackage[T1]{fontenc}
\usepackage[utf8]{inputenc}
\usepackage{microtype}
\usepackage{inconsolata}
\usepackage{graphicx}
\usepackage{booktabs}
\usepackage{placeins}
\usepackage{xcolor}
\usepackage{tikz}

\title{Grammar Concept Annotation at Scale: Deployed Fine-Tuned Small Language Models Outperform Prompted Frontier Models}

\author{Marjan Celikik \\
  Preply \\
  \texttt{marjan.celikik@preply.com} \And
  Ana Peleteiro Ramallo \\
  Preply \\
  \texttt{ana.peleteiro@preply.com} \AND
  Javier Morales \\
  Preply \\
  \texttt{javier.morales@preply.com}}

\begin{document}
\maketitle

\begin{abstract}
Corrective feedback is among the best-evidenced drivers of second-language acquisition,
yet corrections delivered during lessons rarely accumulate into an actionable view of
grammar mastery. Prompted frontier models can provide such a view from learner--tutor
lesson transcripts, but they are costly at scale. We close this gap by fine-tuning
Qwen3.5 small language models (SLMs) on filtered and rebalanced teacher-generated
supervision, then deploying an efficient 0.8B model in an end-to-end grammar mastery tracker for
all English learners on our platform. Internalizing the annotation contract into adapter
weights enables pairing the 0.8B model with a compact matched prompt rather than verbose
instructions. On two human-curated benchmarks, both the deployed 0.8B model and a 4B
reference comparator outperform prompted GPT-5.4 and GPT-5.6 Sol in precision and recall
under nested matching criteria of increasing strictness: concept, evidence span, and
correctness. The deployed 0.8B SLM reduces serving cost by approximately 16$\times$.
A feature-level online experiment shows significant gains in learner engagement ($+15.8\%$)
and key business metrics, including scheduled hours ($+2.1\%$) and GMV from new lessons ($+13.2\%$).
\end{abstract}
 
\section{Introduction}
\label{sec:introduction}

Corrective feedback is one of the best-evidenced drivers of second-language acquisition
\citep{li-2010-effectiveness}. In live tutoring, however, corrections are often given in
the moment rather than retained as structured evidence of progress. This makes it difficult
for learners and tutors to track which grammatical concepts have been mastered, which have
merely been avoided, and where the largest learning gaps remain. An explicit, longitudinal
record bridges this gap: learners see which concepts to practice, while tutors see what to
revisit in future lessons.

We present a grammar mastery tracking system deployed to all English learners on our
platform. It processes speech-to-text transcripts of lesson audio
\footnote{Lesson transcripts are analyzed exclusively from sessions for which we have a lawful basis to do so (see Ethical Considerations).} and aggregates
correct and incorrect concept evidence into overall and per-concept grammar mastery scores
(Figure~\ref{fig:mastery-tracker}). Evidence is
weighted by concept frequency, CEFR level \citep{council-of-europe-2020-cefr}, and a forgetting curve 
that discounts older evidence, consistent with trainable memory models for language learning
\citep{settles-meeder-2016-trainable}. Every piece of evidence is traceable to a
highlighted span in the learner's own sentences; incorrect uses also receive a short explanation
(Figure~\ref{fig:task-contract-example}c).

At the core of the system is the \emph{grammar concept annotation} task that identifies which 
curricular concepts a learner attempts, extracts the corresponding evidence spans, and
judges each use as correct or incorrect. For incorrect uses, it also generates
learner-facing explanations. The model returns these predictions in a single structured pass
per learner turn, capturing both correct mastery and actionable errors.

Our initial product pilot prompted GPT-5.4 for grammar concept annotation in learner
utterances extracted from lesson transcripts. While it achieved the desired annotation
quality, prompting GPT-5.4 at production scale would have far exceeded the feature's serving
budget (Section~\ref{sec:deployment}). Each request repeated the complete concept inventory,
tagging rules, examples, and span conventions. Therefore, the pilot was initially rolled out
to a small fraction of learners. The same pattern is widely reported for enterprise AI:
many products remain in the pilot phase, and only about 5\% reach production
\citep{challapally-etal-2025-genai}.

To make full rollout viable and improve quality, we fine-tuned Qwen3.5 SLMs on
filtered and rebalanced teacher-generated supervision, internalizing the task contract
through demonstrations rather than verbose instructions. We deployed an efficient
0.8B SLM with a compact matched prompt, using a 4B model as a reference comparator to confirm
that aggressive prompt compaction and parameter reduction incur no quality penalty.
On human-curated benchmarks, both models surpass prompted GPT-5.4 and GPT-5.6 Sol.
The contributions of this work are as follows:

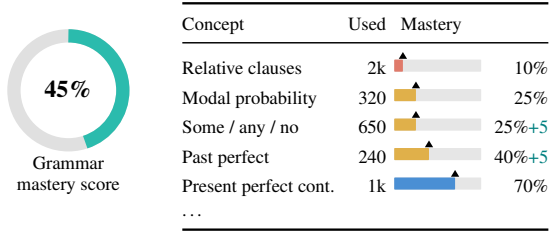
\begin{figure}[t]
  \centering
  \definecolor{masteryteal}{HTML}{2BBBAD}
  \definecolor{masteryred}{HTML}{E07A6A}
  \definecolor{masterygold}{HTML}{E0B04A}
  \definecolor{masteryblue}{HTML}{4A8FD4}
  \newcommand{\gbbar}[2]{%
    \begin{tikzpicture}[baseline=-1pt]
      \fill[black!10, rounded corners=0.3pt] (0,0) rectangle (1.15,0.16);
      \fill[#2, rounded corners=0.3pt] (0,0) rectangle ({1.15*#1/100},0.16);
      \fill[black] ({1.15*#1/100},0.26) -- ++(-0.055,-0.09) -- ++(0.11,0) -- cycle;
    \end{tikzpicture}}
  \begin{minipage}[c]{0.30\columnwidth}
    \centering
    \begin{tikzpicture}
      \draw[line width=5pt, black!12] (0,0) circle (0.72);
      \draw[line width=5pt, masteryteal]
        (90:0.72) arc[start angle=90, end angle=-72, radius=0.72];
      \node[font=\bfseries\small] at (0,0.02) {45\%};
      \node[font=\scriptsize, align=center] at (0,-1.12) {Grammar\\mastery score};
    \end{tikzpicture}
  \end{minipage}%
  \hfill
  \begin{minipage}[c]{0.68\columnwidth}
    \centering
    \scriptsize
    \setlength{\tabcolsep}{2.5pt}
    \renewcommand{\arraystretch}{1.12}
    \begin{tabular}{@{}l r @{\hspace{3pt}} l r@{}}
      \toprule
      Concept & Used & \multicolumn{2}{l}{Mastery} \\
      \midrule
      Relative clauses & 2k & \gbbar{10}{masteryred} & 10\% \\
      Modal probability & 320 & \gbbar{25}{masterygold} & 25\% \\
      Some / any / no & 650 & \gbbar{25}{masterygold} & 25\%{\scriptsize\textcolor{teal}{+5}} \\
      Past perfect & 240 & \gbbar{40}{masterygold} & 40\%{\scriptsize\textcolor{teal}{+5}} \\
      Present perfect cont.\ & 1k & \gbbar{70}{masteryblue} & 70\% \\
      \multicolumn{4}{@{}l}{\ldots} \\
      \bottomrule
    \end{tabular}
  \end{minipage}
  \caption{Product view of the aggregated grammar mastery score (left) and per-concept
    breakdown (right; teal values indicate recent mastery gains).}
  \label{fig:mastery-tracker}
\end{figure}

\begin{enumerate}
\item A scalable end-to-end grammar mastery tracking system, deployed and evaluated in a feature-level
  online experiment showing significant uplift in learner engagement, tutoring hours, and
  gross merchandise value (GMV) from new lessons.
\item A formulation of learner grammar feedback as \emph{grammar concept annotation}
  (Section~\ref{sec:task}), with two human-curated benchmarks and nested criteria for
  concept, exact evidence span, and correctness. On both benchmarks, our deployed 0.8B
  model and 4B reference comparator outperform prompted GPT-5.4 and GPT-5.6 Sol
  at every matching level in precision and recall.
\item A data quality pipeline combining evidence-grounded review, deterministic
  validation, filtering, and synthetic enrichment. Relative to matched 4B and 0.8B
  models trained on raw teacher annotations, the full pipeline raises recall by
  36.9--40.7 percentage points when concept, exact span, and correctness must all
  match, with further gains on rare concepts and incorrect uses.
\item A production serving architecture pairing the task-specialized 0.8B SLM with prompt
  compaction, native vLLM, and autoscaling, reducing serving cost by approximately
  16$\times$ relative to GPT-5.4 based on measured production traffic and load tests.
\end{enumerate}

\section{Grammar Concept Annotation Task}
\label{sec:task}
The production pipeline turns speech-to-text transcripts of lesson audio
into structured evidence for concept-level mastery tracking. It extracts learner
utterances and relevant context, then an upstream LLM produces a corrected sentence
that represents the learner's intended meaning. The annotation model receives the
original utterance and this correction, rather than the full transcript. 

For an input pair $(u_i,c_i)$, where $u_i$ is the learner utterance and $c_i$ its
correction, the system produces a set of annotations
\[
  A(u_i,c_i)=\{(k_j,s_j,y_j,e_j)\}_{j=1}^{m_i},\qquad m_i\ge 0,
\]
where $k_j\in\mathcal{K}$ is one of the $|\mathcal{K}|=34$ grammar concepts,\footnote{The 34 concepts correspond to an initial version of the system; the architecture extends and scales to a much larger number of concepts, with the next iteration containing more than 300 fine-grained concepts.} $s_j$ is an
exact contiguous substring of $u_i$, $y_j\in\{\textsc{correct},\textsc{incorrect}\}$
judges the use of $k_j$ in $s_j$, and $e_j$ is a learner-facing explanation that is
non-empty if and only if $y_j=\textsc{incorrect}$. The model emits this set in a
single pass: one utterance may yield several concepts or none.
Because $s_j$ must be an exact contiguous substring of $u_i$, the same span can be
highlighted in the learner's sentence and compared with gold spans by string identity
at evaluation. 

\begin{figure}[t]
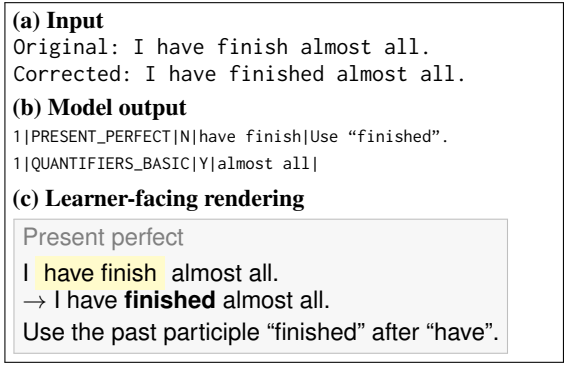

  \centering
  \fbox{\begin{minipage}{0.93\columnwidth}
    \small
    \textbf{(a) Input}\\
    \texttt{Original: I have finish almost all.}\\
    \texttt{Corrected: I have finished almost all.}\\[3pt]
    \textbf{(b) Model output}\\
    {\scriptsize\ttfamily
    1|PRESENT\_PERFECT|N|have finish|Use ``finished''.\\
    1|QUANTIFIERS\_BASIC|Y|almost all|}\\[4pt]
    \textbf{(c) Learner-facing rendering}\\[2pt]
    \fcolorbox{black!25}{black!3}{%
      \begin{minipage}{0.88\columnwidth}
        \textsf{\textcolor{gray}{Present perfect}}\\[1pt]
        \textsf{I \colorbox{yellow!25}{have finish} almost all.}\\
        \textsf{$\rightarrow$ I have \textbf{finished} almost all.}\\[2pt]
        \textsf{Use the past participle ``finished'' after ``have''.}
      \end{minipage}}
  \end{minipage}}
  \caption{Illustrative synthetic example. (a) Learner utterance and upstream correction.
    (b) Model output: sentence index, concept, correctness (Y/N), exact source span, and
    explanation for incorrect use only. (c) Learner-facing rendering of one incorrect
    annotation; correct annotations contribute to mastery without appearing as feedback.}
  \label{fig:task-contract-example}
\end{figure}

Figure~\ref{fig:task-contract-example} shows an example input, model output, and
learner-facing rendering. Appendix Table~\ref{tab:task-schema} states the same
contract field by field. The runtime representation is one pipe-delimited
annotation per line. The format was chosen during the product pilot to balance
quality and cost.

The nested structure of the tuple defines the evaluation hierarchy in
Section~\ref{sec:evaluation}: matching on $k_j$, on $(k_j,s_j)$, and on
$(k_j,s_j,y_j)$ yields progressively stricter criteria, while $e_j$ is assessed
separately because explanations are free text.

\section{Data and Quality Pipeline}
\label{sec:data}

\subsection{Training Data Collection, Cleaning, and Enrichment}
\label{sec:data-pipeline}
Directly fine-tuning on raw production logs is hindered by teacher annotation noise,
span boundary drift, and severe class imbalance favoring frequent, error-free structures.
We address these challenges through a three-stage curation pipeline:
(i)~evidence-grounded review and programmatic validation of production annotations;
(ii)~structured synthetic generation to rebalance rare concepts and pedagogical errors; and
(iii)~corpus consolidation, deduplicating across sources (1.4\% overlap) and strictly
removing any utterance present in either golden benchmark.
The resulting training corpus contains approximately 187k utterance groups (learner
utterance, correction, and annotations), consolidated from 100k reviewed production records
and 90k synthetic records prior to deduplication.

\paragraph{Production data and cleaning.}
Each production group pairs a learner utterance and upstream correction with span-level
annotations from the product pilot. We sample 50k groups uniformly to preserve
production traffic distributions, and 50k via heuristic sampling prioritizing utterances
suspected of incomplete or excessive annotations (anomalous counts relative to length,
or low span coverage). To clean these records, we adopt a hybrid neuro-symbolic strategy:
deterministic linguistic anchors extracted via spaCy \citep{honnibal-etal-2020-spacy}
are converted into a plain-text inventory of grammatical elements (articles, pronouns,
prepositions, modals, verb forms, and tense patterns), alongside hand-crafted candidate rules.
In the prompt to GPT-5.4, these cues are combined with word-level edit differences from the
upstream correction (pinpointing where the learner diverged from target grammar) and the
pilot annotations. Rather than filtering alone, GPT-5.4 performs a full review and
re-annotation pass (retaining, correcting, removing, or adding tuples). A deterministic
post-validator then enforces strict contract invariants: rejecting non-substring spans,
pruning invalid labels, and enforcing explanation consistency (Appendix~\ref{sec:pipeline-details}).

\paragraph{Synthetic enrichment.}
Unconstrained synthetic generation often hallucinates labels or perpetuates natural class
skews. We instead decouple syntactic scaffolding from lexical realization using structural
\emph{backbones}---sequences of concept labels and correctness flags harvested from quality-filtered
production utterances. To address long-tail sparsity, we stratify backbone sampling across
three 30k subsets: uniform sampling, error-weighted sampling (favoring templates with natural
mistakes), and smoothed inverse concept-frequency sampling (boosting rare curricular targets).
Unlike post-hoc token corruption, which produces disfluent or unnatural errors, we inject
pedagogical errors directly into the generation plan: with low probability, a correct slot is assigned
an error type (within-concept misuse, omission, or confusable structure substitution).
Prompted with the backbone and matching production exemplars, GPT-5.4 realizes the plan into
learner-style sentences, exact spans, and targeted explanations. Deterministic post-validation
then enforces span substring constraints and label alignment (Appendix~\ref{sec:pipeline-details}).

\subsection{Golden Evaluation Benchmarks}
Evaluating an educational tracking system requires measuring both overall operational
accuracy and robustness on critical failure modes. We curate two independent human-adjudicated
benchmarks over the 34 concepts (Table~\ref{tab:golden-stats}). \emph{Golden} (689 utterances)
mirrors the natural production distribution, providing an unskewed evaluation of daily traffic.
In contrast, \emph{Golden2} (2,977 utterances, constructed via model-assisted
candidate retrieval followed by double-annotator curation and adjudication) is an
error-enriched diagnostic benchmark (95.6\% error-bearing utterances vs.\ 50.1\% in Golden).
Golden2 stress-tests the system against the product's costliest failure mode---failing to detect
a genuine learner error---while providing ample support to evaluate explanation fidelity.
In both benchmarks, two annotators worked independently and a third adjudicated disagreements.

\begin{table}[t]
  \centering
  \scriptsize
  \setlength{\tabcolsep}{3pt}
  \begin{tabular*}{\columnwidth}{@{\extracolsep{\fill}}lrr}
    \toprule
    Statistic & Golden & Golden2 \\
    \midrule
    Utterances & 689 & 2,977 \\
    Grammar concepts & 34 & 34 \\
    Concept annotations & 3,598 & 17,400 \\
    Annotations / utterance & 5.22 & 5.84 \\
    Annotations marked incorrect & 592 (16.5\%) & 6,075 (34.9\%) \\
    Utterances with $\geq$1 error & 345 (50.1\%) & 2,845 (95.6\%) \\
    \bottomrule
  \end{tabular*}
  \caption{Composition of the two golden evaluation benchmarks under the common
    34-concept evaluation scope. Annotation counts are computed after scope filtering.}
  \label{tab:golden-stats}
\end{table}

Learner ASR transcripts, annotations, and internal prompts are proprietary and
privacy-sensitive, so we report aggregate statistics and methodological details but do
not release records. Appendix~\ref{sec:synthetic-probe} reports an additional
teacher-synthetic distribution-shift probe.

\section{Fine-Tuning Method}
\label{sec:method}
We adapted Qwen3.5 \citep{qwen-team-2026-qwen35} and Qwen3
\citep{yang-etal-2025-qwen3} base models with completion-only LoRA
\citep{hu-etal-2022-lora}. At training and serving time, packing five
utterance--correction pairs into one conversation (Section~\ref{sec:deployment})
balanced throughput and memory; scaling further yielded diminishing returns while
increasing activation memory. We used a 90/10 train--validation split and discarded
examples exceeding the maximum sequence length rather than truncating targets.
Checkpoints were selected by minimum teacher-forced validation loss evaluated
four times per epoch with early-stopping patience of three evaluations
(Appendix~\ref{sec:training-convergence}).
	
\paragraph{Prompt internalization and model sizing.}
Our primary deployment target is an efficient, sub-1B parameter model
(Qwen3.5-0.8B) capable of serving high-concurrency traffic on single-GPU commodity
instances (\texttt{g6.2xlarge}). In the production pilot, GPT-5.4 required a
3,810-token prompt detailing all 34 concepts, edge cases, and examples; at scale,
repeatedly ingesting this verbose prompt induces severe prefill latency and dominates
KV-cache memory in vLLM \citep{kwon2023efficient}, restricting concurrency.
However, task fine-tuning enables \emph{prompt internalization}
\citep{zou-etal-2024-promptintern,shin-etal-2025-generative}: the model absorbs
concept definitions and span conventions directly into adapter weights from demonstrations.
We therefore designed a compact prompt (375 Qwen3.5 tokens, a 90.2\% reduction)
specifying only concept names, the schema contract,
and brief rules, shrinking the maximum sequence length from 7,000 to 2,500 tokens (used for
both 0.8B and 0.6B; Appendix~\ref{sec:training-convergence}). To provide sufficient capacity for these rules,
the 0.8B model used a higher adapter rank ($r=128$ across all linear layers vs.\ $r=16$ for 4B;
Appendix Table~\ref{tab:training-config}). The larger Qwen3.5-4B model retained the detailed
prompt strictly as a reference comparator to confirm that prompt compaction and parameter
reduction incur no quality penalty. Differences between sizes thus reflect distinct
operational recipes rather than an isolated parameter ablation.

\section{Evaluation Methodology}
\label{sec:evaluation}

\begin{table*}[t]
  \centering
  \scriptsize
  \setlength{\tabcolsep}{2pt}
  \begin{tabular*}{\textwidth}{@{\extracolsep{\fill}}lrrrrrrrrrrrr}
    \toprule
    & \multicolumn{2}{c}{GPT-5.4}
      & \multicolumn{2}{c}{GPT-5.6 Sol}
      & \multicolumn{4}{c}{Qwen3.5-4B}
      & \multicolumn{4}{c}{Qwen3.5-0.8B} \\
    \cmidrule(lr){2-3}\cmidrule(lr){4-5}\cmidrule(lr){6-9}\cmidrule(lr){10-13}
    & \multicolumn{2}{c}{Prompted (prod.)}
      & \multicolumn{2}{c}{Prompted}
      & \multicolumn{2}{c}{Base}
      & \multicolumn{2}{c}{Fine-tuned}
      & \multicolumn{2}{c}{Base}
      & \multicolumn{2}{c}{Fine-tuned$^\dagger$} \\
    \cmidrule(lr){2-3}\cmidrule(lr){4-5}\cmidrule(lr){6-7}
    \cmidrule(lr){8-9}\cmidrule(lr){10-11}\cmidrule(lr){12-13}
    Criterion & \shortstack{P\\(\%)} & \shortstack{R\\(\%)}
      & \shortstack{P\\(\%)} & \shortstack{R\\(\%)}
      & \shortstack{P\\(\%)} & \shortstack{R\\(\%)}
      & \shortstack{P\\(\%)} & \shortstack{R\\(\%)}
      & \shortstack{P\\(\%)} & \shortstack{R\\(\%)}
      & \shortstack{P\\(\%)} & \shortstack{R\\(\%)} \\
    \midrule
    \multicolumn{13}{l}{\emph{Golden}} \\
    L1: concept & 82.34 & 76.35 & 87.69 & 89.66 & 6.12 & 33.69 & 91.26 & 93.75
      & 0.20 & 4.53 & \textbf{91.68} & \textbf{94.36} \\
    L2: + span & 73.20 & 67.87 & 79.59 & 81.38 & 3.72 & 20.46 & 85.82 & 88.16
      & 0.02 & 0.47 & \textbf{86.88} & \textbf{89.41} \\
    L3: + correctness & 67.81 & 62.87 & 74.37 & 76.04 & 3.01 & 16.56 & 81.28 & 83.49
      & 0.01 & 0.19 & \textbf{82.74} & \textbf{85.16} \\
    \addlinespace
    \multicolumn{13}{l}{\emph{Golden2}} \\
    L1: concept & 84.89 & 74.28 & 89.65 & 86.55 & 9.04 & 31.31 & 92.53 & 90.30
      & 0.20 & 4.10 & \textbf{93.48} & \textbf{91.59} \\
    L2: + span & 73.17 & 64.02 & 78.98 & 76.25 & 4.35 & 15.07 & 86.09 & 84.02
      & 0.02 & 0.47 & \textbf{87.04} & \textbf{85.28} \\
    L3: + correctness & 67.04 & 58.66 & 74.97 & 72.39 & 3.10 & 10.73 & 80.84 & 78.89
      & 0.01 & 0.21 & \textbf{82.55} & \textbf{80.88} \\
    \bottomrule
  \end{tabular*}
  \caption{Micro precision and recall (\%) on Golden and Golden2. GPT-5.4 is the
    production prompted model; GPT-5.6 Sol is a quality-only comparator. Bold marks the
    best value in each row. Base denotes untuned Qwen3.5 models; fine-tuned denotes
    selected full-pipeline checkpoints (raw-supervision controls in
    Table~\ref{tab:pipeline-ablation}). The two fine-tuned models use independently tuned
    recipes; the 0.8B checkpoint, marked $^\dagger$, combines a compact prompt with
    separately tuned hyperparameters, so differences between sizes are not a controlled
    comparison. Bootstrap 95\% confidence intervals appear in Appendix
    Table~\ref{tab:annotation-ci}; key-macro results in Appendix
    Table~\ref{tab:annotation-macro}. All fine-tuned Qwen--GPT-5.4 differences are
    significant under paired randomization tests ($p<0.0001$).}
  \label{tab:annotation-results}
\end{table*}

\paragraph{Annotation evaluation.}
Each system predicts a set $\hat{A}(u_i,c_i)$ over the 34-concept inventory, scored
against gold $A(u_i,c_i)$ from Section~\ref{sec:task}. We benchmark against two
reference classes: frontier models establish operational parity (production GPT-5.4)
and a strong quality comparator (GPT-5.6 Sol), while base-model comparisons isolate
task-adaptation gains against untuned counterparts. Unparseable outputs are recorded
as format-adherence failures (parse errors) and treated as empty predictions,
measuring syntactic reliability under autoregressive generation.

We score the nested criteria of Section~\ref{sec:task} at L1 ($k_j$), L2 ($(k_j,s_j)$),
and L3 ($(k_j,s_j,y_j)$), reporting micro precision and recall across utterances
(key-macro averages in Appendix~\ref{sec:macro-results}). Test-set uncertainty is estimated
via percentile 95\% confidence intervals from 20,000 paired nonparametric cluster
bootstrap resamples of packed five-utterance inference batches, preserving output
dependencies. Differences from GPT-5.4 are assessed via two-sided paired randomization
tests with 20,000 permutations \citep{koehn-2004-statistical,dror-etal-2018-hitchhikers}
(Appendix~\ref{sec:stat-details}).

\paragraph{Explanation evaluation.}
Because surface n-gram metrics correlate poorly with pedagogical feedback quality,
we adopt an LLM-as-a-judge framework \citep{zheng-etal-2023-judging,kobayashi-etal-2024-large}:
GPT-5.5 evaluates explanation acceptability given the utterance, span, reference
correction, and concept definition (rubric in Appendix~\ref{sec:llmj-rubric}). To counter
potential model leniency, a human reviewer audits the judgments, and an independent
second reviewer reconciles disagreements.

\paragraph{Pipeline ablation.}
To isolate pipeline contributions, we trained raw-supervision controls using unreviewed
GPT-5.4 annotations with identical prompts, splits, hyperparameters, and matched
optimizer-update budgets. Because model sizes use independently tuned recipes, each
control is compared only against its same-size counterpart. To evaluate how well the
pipeline addresses specific distribution skews, we report equal-weight L3 recall across
all 34 concepts (Mean-34; curricular balance), micro L3 recall restricted to gold incorrect
uses (Incorrect; error sensitivity), and equal-weight L3 recall on \emph{Rare-9} (the nine
least frequent concepts in raw training data, validating synthetic tail-rebalancing;
Appendix~\ref{sec:rare-nine}).

\section{Results}
\label{sec:results}

\paragraph{Annotation accuracy and specialization.}
Table~\ref{tab:annotation-results} reports precision and recall on both benchmarks. Both
fine-tuned models outperform GPT-5.6 Sol across all criteria, with the largest uplift
under strict L3 matching (concept, exact span, and correctness).

\paragraph{Base-model controls and format adherence.}
Untuned 4B and 0.8B models fail to parse on 26.9--32.8\% and 75.3--92.0\% of examples,
respectively, demonstrating a wide adaptation gap. An earlier-generation Qwen3-0.6B
checkpoint trained with the 0.8B recipe produced unparseable output on 57.4\% of Golden2
versus 0.03\% for 0.8B despite smooth validation convergence (generational differences
prevent attributing this gap solely to parameter count;
Appendix~\ref{sec:training-convergence}).

\begin{table}[t]
  \centering
  \scriptsize
  \setlength{\tabcolsep}{2.2pt}
  \begin{tabular*}{\columnwidth}{@{\extracolsep{\fill}}llrrrrr}
    \toprule
    Model & Training data & \shortstack{L3\\P}
      & \shortstack{L3\\R} & \shortstack{Mean-34\\L3 R}
      & \shortstack{Rare-9\\L3 R} & \shortstack{Incorrect\\L3 R} \\
    \midrule
    \multicolumn{7}{l}{\emph{Golden2}} \\
    Qwen3.5-4B & Raw supervision & 52.47 & 41.98 & 34.83 & 22.04 & 29.35 \\
                & Full pipeline & 80.84 & 78.89 & 69.44 & \textbf{67.35} & 67.56 \\
    \addlinespace[1pt]
    Qwen3.5-0.8B & Raw supervision & 52.62 & 40.93 & 35.25 & 21.05 & 29.68 \\
                  & Full pipeline & \textbf{82.55} & \textbf{80.88} & \textbf{72.64} & 66.86 & \textbf{69.37} \\
    \addlinespace
    \multicolumn{7}{l}{\emph{Golden}} \\
    Qwen3.5-4B & Raw supervision & 57.34 & 46.16 & 44.70 & 38.67 & 22.64 \\
                & Full pipeline & 81.28 & 83.49 & 75.99 & 70.96 & 52.70 \\
    \addlinespace[1pt]
    Qwen3.5-0.8B & Raw supervision & 56.59 & 44.50 & 44.21 & 40.55 & 20.61 \\
                  & Full pipeline & \textbf{82.74} & \textbf{85.16} & \textbf{76.89} & \textbf{71.97} & \textbf{55.57} \\
    \bottomrule
  \end{tabular*}
  \caption{Pipeline ablation (\%) on Golden2 and Golden. Bold marks the best value in
    each column within each benchmark. Mean-34 is equal-weight recall
    over all concepts; Rare-9 is equal-weight recall over the nine concepts fixed from
    the raw training split; Incorrect restricts recall to gold incorrect-use
    annotations. Full-pipeline rows are the selected checkpoints in
    Table~\ref{tab:annotation-results}.}
  \label{tab:pipeline-ablation}
\end{table}

\paragraph{Data-pipeline ablation.}
Matched within-size controls isolate the combined supervision pipeline
(Table~\ref{tab:pipeline-ablation}). On Golden2 it raises L3 precision by 28.37 (4B) and
29.93 (0.8B) points and L3 recall by 36.91 (95\% CI [35.93, 37.87]) and 39.95
([39.01, 40.89]) points; on Golden, L3 recall rises by 37.33 ([35.15, 39.48]) and
40.66 ([38.44, 42.81]) points. Gains extend across all subsets, notably Rare-9 (up
45.31 and 45.81 points on Golden2). On Golden2, raw-supervision recall drops by
23.64--23.78 points from L1 to exact-span L2, compared to 6.28--6.31 points for the full
pipeline, underscoring an acute exact-span degradation under raw supervision.

\paragraph{Explanation quality.}
On Golden2, GPT-5.5 judged 99.06\% of GPT-5.4 explanations acceptable versus 94.91\%
(4B) and 93.88\% (0.8B). However, following human audit and reconciliation, all three
systems achieved similar acceptable rates ($\approx$95\%) with no significant differences.
This disparity indicates that the uncalibrated LLM judge favored its frontier sibling,
underscoring the necessity of human calibration.

\paragraph{Online product impact.}
We deployed the complete system with Qwen3.5-0.8B (Section~\ref{sec:deployment}) in a
35-day feature-level A/B experiment randomized at the learner level with 50\% exposure.
At the fixed pre-scaling snapshot (Table~\ref{tab:online-experiment}), learner engagement
with progress and all three business metrics improved significantly. These gains apply
to the end-to-end product feature rather than the SLM in isolation.

\begin{table}[t]
  \centering
  \footnotesize
  \setlength{\tabcolsep}{3pt}
  \begin{tabular*}{\columnwidth}{@{\extracolsep{\fill}}lrr}
    \toprule
    Metric & Effect & $p$ \\
    \midrule
    Engagement with their progress & \textbf{+15.80\%} & $<0.001$ \\
    Scheduled hours (CUPED) & \textbf{+2.06\%} & $0.021$ \\
    Confirmed hours from new tutoring & \textbf{+10.87\%} & $0.020$ \\
    GMV from new lessons & \textbf{+13.20\%} & $0.039$ \\
    \bottomrule
  \end{tabular*}
  \caption{Relative effects in the online experiment at the fixed pre-scaling evaluation
    snapshot (all corresponding 95\% CIs exclude zero; bold marks $p < 0.05$).}
  \label{tab:online-experiment}
\end{table}

\section{Deployment and Operational Considerations}
\label{sec:deployment}

Our production serving architecture prioritizes high concurrent throughput and cost-efficient
horizontal scaling. We deploy Qwen3.5-0.8B with vLLM \citep{kwon2023efficient} on autoscaled,
single-GPU \texttt{g6.2xlarge} SageMaker instances that allow fine-grained capacity scaling.
Before inference, the application packs prepared utterance--correction pairs into five-pair
requests to an internal inference service. Requests are sequential within each transcript chunk,
and vLLM batches concurrent requests arriving at each replica. The service applies the compact
prompt, parses the structured output, and returns annotations
aligned to input indices.
Validation or service failures route work to GPT-5.4 (topology in Appendix
Figure~\ref{fig:serving-setup}; autoscaling and load testing in
Appendix~\ref{sec:autoscaling-validation}).

The warmed load test exercised this packed-request path: 8 \texttt{g6.2xlarge}
instances sustained throughput above the observed post-rollout peak.
During a 31-day production APM snapshot, p95 latency was 4.3 seconds. Combining measured GPT-5.4 production cost, observed demand, load-tested
SLM capacity, and public SageMaker prices, recurring detection cost per lesson is 6.2\% of the
GPT-5.4 baseline---an approximately 16$\times$ reduction (Appendix~\ref{sec:cost-model}).

\section{Related Work}
\label{sec:related-work}
\paragraph{Task formulation.}
Grammatical error detection and correction localize or repair deviations
\citep{rei-yannakoudakis-2016-compositional,bryant-etal-2023-grammatical}, label edits, or
generate learner feedback
\citep{bryant-etal-2017-automatic,nagata-2019-toward,liang-etal-2023-chatback,song-etal-2024-gee}.
Curriculum-grounded methods often target predefined activities or dialogue contrasts
\citep{quixal-etal-2021-automatic,akef-etal-2026-llm,ribeiro-flucht-etal-2026-dialogue}.
In contrast, we identify concepts from a 34-concept inventory in unconstrained transcribed
speech turns, jointly predicting evidence spans, correctness, and explanations.

\paragraph{Teacher-generated and synthetic supervision.}
Prior work generates synthetic GEC data via tagged corruption or LLM prompting and adapts
distributions via type-aware augmentation or reweighting
\citep{stahlberg-kumar-2021-synthetic,li-lan-2025-large,vainikko-etal-2025-paragraph,
zhang-etal-2025-synthesizing}. Our pipeline instead filters production annotations and samples
concept--correctness backbones to rebalance the curriculum, benchmarked against matched
raw-supervision controls.

\paragraph{SLM specialization and deployment.}
Prior work internalizes prompts
\citep{zou-etal-2024-promptintern,shin-etal-2025-generative}, distills teacher data into
lightweight models \citep{wang-etal-2025-distilqwen2}, or supplies LLM explanations to
SLM GEC systems \citep{li-etal-2025-rethinking}. We combine prompt internalization,
a compact schema prompt, and completion-only LoRA in a deployed sub-1B SLM, jointly
evaluating accuracy, serving cost, and online impact.

\section{Conclusion}
\label{sec:conclusion}
We presented a deployed grammar tracker jointly predicting curricular concepts, exact
evidence spans, correctness, and explanations for errors. Fine-tuned Qwen3.5 models
outperformed GPT-5.4 and GPT-5.6 Sol across all matching levels on two human-curated
benchmarks. Ablations confirmed our data pipeline yields 36.9--40.7 points of strict L3
recall over raw teacher supervision. The deployed 0.8B model cut serving cost by
$\approx$16$\times$, while an online experiment demonstrated significant engagement and
business gains, showing that data quality, prompt internalization, and efficient serving
make accurate feedback viable at scale.

\clearpage
\section*{Limitations}
\paragraph{Modeling and training controls.}
Our pipeline ablation benchmarks raw supervision against the combined pipeline (review,
deterministic validation, filtering, and synthetic enrichment), but does not isolate the
individual contribution of each component. Furthermore, we report results from a single
training run per configuration and do not estimate across-seed variance, meaning fine-grained
metric differences should not be over-interpreted. Finally, because the 4B and 0.8B models
rely on independently tuned training recipes and prompt lengths, and the 0.6B comparison
belongs to an earlier Qwen generation, differences across model sizes reflect distinct
operational configurations rather than an isolated parameter ablation.

\paragraph{Operational and evaluation scope.}
The online A/B experiment evaluates the end-to-end product feature rather than the SLM or
serving infrastructure in isolation. Similarly, our serving cost model isolates recurring
grammar-detection compute and excludes upstream transcription, correction, and fallback
calls to frontier APIs. Regarding explanation evaluation, while human audit reconciled
acceptability rates, the leniency gap in the uncalibrated LLM judge demonstrates the need
for domain-specific judge calibration; moreover, explanations were evaluated against gold
rubrics rather than through direct perceptual studies with teachers or learners.

\paragraph{Data scope, speech, and supervision biases.}
While our deployed checkpoints target an English-language curriculum of 34 concepts, the
underlying pipeline---curated supervision, prompt internalization, and lightweight
serving---is framework-agnostic and designed to scale to a
substantially broader curriculum. Because learner turns are transcribed via automatic
speech recognition (ASR), transcription errors can cascade into upstream correction and
downstream concept tracking; we do not isolate robustness to ASR noise. Finally, because
both reviewed production supervision and synthetic sentences derive from GPT-5.4,
systematic teacher biases may persist in ways our benchmarks cannot fully reveal.

\section*{Ethical Considerations}
We analyze lesson transcripts only from sessions for which we have a lawful basis to do so. 
Learner utterances, annotations, and internal prompts are not released; the illustrative 
example in Figure~\ref{fig:task-contract-example} is synthetic. The paper reports 
aggregate evaluation and online-experiment results only.


\bibliography{references}

\clearpage
\appendix
\makeatletter
\setlength{\@fptop}{0pt}
\setlength{\@fpsep}{12pt}
\setlength{\@fpbot}{0pt plus 1fil}
\makeatother
\section{Training Configurations and Convergence}
\label{sec:training-convergence}

All runs used bfloat16, gradient checkpointing \citep{chen-etal-2016-sublinear}, DeepSpeed
ZeRO-2 \citep{rajbhandari-etal-2020-zero}, completion-only loss, and long-sample
filtering. Validation was performed four times per epoch with early-stopping patience of
three evaluations. Table~\ref{tab:training-config} records the selected run
configurations, and Table~\ref{tab:model-convergence} reports training convergence and
structured-output reliability for the three fine-tuned models.

\begin{table}[!htbp]
  \centering
  \footnotesize
  \setlength{\tabcolsep}{3pt}
  \begin{tabular*}{\columnwidth}{@{\extracolsep{\fill}}lrrr}
    \toprule
    Setting & \shortstack{Qwen3.5\\4B} & \shortstack{Qwen3.5\\0.8B}
      & \shortstack{Qwen3\\0.6B} \\
    \midrule
    Prompt design & Detailed & Compact & Compact \\
    Max.\ sequence length & 7,000 & 2,500 & 2,500 \\
    LoRA $r/\alpha$ & 16 / 32 & 128 / 64 & 128 / 64 \\
    LoRA dropout & --- & 0.05 & 0.05 \\
    LoRA targets & --- & All linear & All linear \\
    Learning rate & $2\times10^{-5}$ & $1\times10^{-4}$ & $1\times10^{-4}$ \\
    LR schedule & --- & Cosine & Cosine \\
    Warm-up & --- & 3\% & 3\% \\
    Batch / device & 1 & 1 & 1 \\
    Grad.\ accumulation & 16 & 32 & 32 \\
    Effective batch & 16 & 256 & 256 \\
    Selected epoch & 15.00 & 7.50 & 8.70 \\
    \bottomrule
  \end{tabular*}
  \caption{Recorded fine-tuning configurations for Qwen3.5-4B, Qwen3.5-0.8B,
    and Qwen3-0.6B. Effective batch size (256 for the compact runs) is the product of
    per-device batch size (1), gradient accumulation steps (32), and 8 distributed GPU workers
    ($1 \times 32 \times 8 = 256$); the 4B run used 1 worker with accumulation 16 ($1 \times 16 \times 1 = 16$).
    Dashes indicate settings not retained in the available run record.}
  \label{tab:training-config}
\end{table}

\begin{table}[!htbp]
  \centering
  \footnotesize
  \setlength{\tabcolsep}{3pt}
  \begin{tabular*}{\columnwidth}{@{\extracolsep{\fill}}lrrr}
    \toprule
    Model & Val.\ loss & Epoch & Parse errors \\
    \midrule
    Qwen3.5-4B$^\dagger$ & \textbf{0.2201} & 15.00 & \textbf{0 / 2,977 (0.00\%)} \\
    Qwen3.5-0.8B & \textbf{0.2245} & 7.50 & \textbf{1 / 2,977 (0.03\%)} \\
    Qwen3-0.6B & 0.2885 & 8.70 & 1,710 / 2,977 (57.44\%) \\
    \bottomrule
  \end{tabular*}
  \caption{Minimum teacher-forced loss on each run's training-validation split and
    structured-output parse errors on Golden2. Bold highlights format adherence
    ($\le 0.03\%$ parse errors) and the lowest validation loss within each distinct
    recipe (0.2201 for detailed 4B; 0.2245 for compact 0.8B). The 0.8B and 0.6B runs
    shared prompt and optimization hyperparameters but selected different early-stopping
    checkpoints. $^\dagger$ The 4B checkpoint used a different prompt, training target,
    and schedule; its loss is descriptive rather than a controlled cross-model comparison.}
  \label{tab:model-convergence}
\end{table}

The matched-recipe comparison between the 0.8B and 0.6B runs reveals a failure mode not
visible in the loss curve. Although the 0.8B and 0.6B runs shared prompt and optimization
hyperparameters, the 0.6B model converged smoothly to a minimum validation loss 28.5\%
higher, and 57.4\% of its evaluation outputs failed schema validation (parse errors),
compared with 0.03\% for the 0.8B model: token-level optimization did not guarantee
reliable autoregressive generation of the required structure. Because the models also
belong to different Qwen generations, this comparison does not isolate parameter count
as the sole cause.

\FloatBarrier
\section{Data Pipeline Details}
\label{sec:pipeline-details}
\paragraph{Error-prioritized sampling.}
The heuristic used to select the second 50k production groups flags annotation
density (annotations per non-noun word) when it is far from one, or when
annotated-span coverage is low. Flagged utterances are ranked by how far density is from one
and by how much coverage is missing, with that score dampened for very short utterances.

\paragraph{Deterministic evidence and validation.}
Beyond part-of-speech tags and rule-based construction detection, coherence checks and
high-confidence rules add warnings and candidate concepts to the review checklist. In
addition to the checks listed in Section~\ref{sec:data-pipeline}, the validator resolves
disfluent repetitions omitted by the correction before checking span containment, and
falls back to the pilot annotations if all proposed tuples are rejected.

\paragraph{Review calls.}
Review calls query GPT-5.4 on one utterance at a time.

\paragraph{Synthetic backbone generation.}
Seed backbones are harvested from production records by stripping lexical tokens from
annotated utterances, after discarding degenerate sequences (e.g., word repetitions or
comma-heavy enumerations). The 90k synthetic corpus is produced across three 30k strata:
(i)~\emph{uniform sampling} from production backbones;
(ii)~\emph{error-weighted sampling}, which upweights frequent templates that contain more
natural errors, with diminishing returns as the error count grows; and
(iii)~\emph{smoothed inverse-frequency sampling}, which boosts rare curricular targets.

\paragraph{Pedagogical error injection and synthesis.}
Each originally correct slot is mutated with low probability into one of three error kinds:
(a)~\emph{same-label misuse} (e.g., wrong verb inflection);
(b)~\emph{omission} (dropping an obligatory article or preposition); or
(c)~\emph{confusion swap}, where the slot is temporarily projected to a confusable substitute
label (e.g., present continuous instead of present perfect continuous) and paired with discourse
cues (e.g., duration adverbials) that make the substitute form erroneous in context.
GPT-5.4 realizes each projected sequence into learner-style sentences. Post-processing restores
confusion-swap slots to the intended label, coerces incorrect spans to the planned intent,
drops sentences with out-of-scope labels, and enforces substring validity.

\FloatBarrier
\twocolumn[{
\begin{minipage}{\textwidth}
  \section{Extended Experimental Results and Statistical Methodology}
  \label{sec:extended-results}

  \subsection{Confidence Intervals for Annotation Results}
  \label{sec:annotation-ci}

  Table~\ref{tab:annotation-ci} provides the complete percentile 95\% confidence
  intervals corresponding to the primary evaluation in Table~\ref{tab:annotation-results},
  derived from 20,000 paired cluster bootstrap resamples.

  \centering
  \scriptsize
  \setlength{\tabcolsep}{4pt}
  \renewcommand{\arraystretch}{1.15}
  \begin{tabular*}{\textwidth}{@{\extracolsep{\fill}}lcccccc}
    \toprule
    Model & L1 P & L1 R & L2 P & L2 R & L3 P & L3 R \\
    \midrule
    \multicolumn{7}{l}{\emph{Golden}} \\
    GPT-5.4 & [80.28,84.19] & [74.36,78.29] & [70.99,75.22]
      & [65.69,70.00] & [65.62,69.87] & [60.68,65.07] \\
    \addlinespace[1pt]
    GPT-5.6 Sol & [86.58,88.76] & [88.29,90.94] & [78.31,80.87]
      & [79.88,82.83] & [72.74,75.98] & [74.28,77.81] \\
    \addlinespace[1pt]
    Qwen3.5-4B Base & [4.47,8.73] & [30.67,36.71] & [2.68,5.37]
      & [18.19,22.76] & [2.15,4.38] & [14.44,18.74] \\
    \addlinespace[1pt]
    Qwen3.5-4B FT & [90.22,92.28] & [92.81,94.66] & [84.51,87.12]
      & [86.95,89.34] & [79.86,82.71] & [82.06,84.93] \\
    \addlinespace[1pt]
    Qwen3.5-0.8B Base & [0.17,0.24] & [3.86,5.22] & [0.01,0.03]
      & [0.25,0.74] & [0.00,0.02] & [0.06,0.37] \\
    \addlinespace[1pt]
    Qwen3.5-0.8B FT & \textbf{[90.63,92.69]} & \textbf{[93.45,95.23]} & \textbf{[85.72,88.01]}
      & \textbf{[88.28,90.53]} & \textbf{[81.44,84.03]} & \textbf{[83.85,86.48]} \\
    \addlinespace
    \multicolumn{7}{l}{\emph{Golden2}} \\
    GPT-5.4 & [84.26,85.52] & [73.52,75.03] & [72.35,73.99]
      & [63.17,64.88] & [66.18,67.91] & [57.79,59.54] \\
    \addlinespace[1pt]
    GPT-5.6 Sol & [89.15,90.14] & [86.02,87.09] & [78.29,79.66]
      & [75.57,76.95] & [74.24,75.72] & [71.67,73.11] \\
    \addlinespace[1pt]
    Qwen3.5-4B Base & [7.88,10.45] & [30.06,32.56] & [3.77,5.06]
      & [14.24,15.93] & [2.67,3.61] & [10.04,11.44] \\
    \addlinespace[1pt]
    Qwen3.5-4B FT & [92.11,92.95] & [89.87,90.73] & [85.54,86.64]
      & [83.46,84.57] & [80.16,81.51] & [78.21,79.57] \\
    \addlinespace[1pt]
    Qwen3.5-0.8B Base & [0.18,0.22] & [3.81,4.40] & [0.02,0.03]
      & [0.36,0.59] & [0.01,0.01] & [0.14,0.29] \\
    \addlinespace[1pt]
    Qwen3.5-0.8B FT & \textbf{[93.09,93.85]} & \textbf{[91.16,92.00]} & \textbf{[86.51,87.57]}
      & \textbf{[84.71,85.85]} & \textbf{[81.91,83.20]} & \textbf{[80.21,81.55]} \\
    \bottomrule
  \end{tabular*}
  \captionof{table}{Percentile 95\% confidence intervals for micro precision and recall
    in Table~\ref{tab:annotation-results}, based on 20,000 paired nonparametric cluster
    bootstrap resamples of packed five-utterance inference batches. Surplus generated
    utterances remain assigned to their originating batch. Bold marks the best interval
    in each column within each benchmark. Base and FT denote untuned and
    fine-tuned models, respectively. All values are percentages; all GPT-5.4 and GPT-5.6
    Sol outputs parsed successfully.}
  \label{tab:annotation-ci}
\end{minipage}
}]

\subsection{Key-Macro Annotation Results}
\label{sec:macro-results}

Table~\ref{tab:annotation-macro} presents key-macro-averaged precision and recall, assigning
equal weight to each distinct target key at each evaluation level (target concepts at L1,
concept--span pairs at L2, and concept--span--correctness triples at L3), complementing
the class-level Mean-34 recall metric in Table~\ref{tab:pipeline-ablation}. Macro recall
and precision are computed as the unweighted mean of per-key scores across all distinct
target and predicted keys, respectively.

\begin{table}[!htbp]
  \centering
  \scriptsize
  \setlength{\tabcolsep}{2pt}
  \begin{tabular*}{\columnwidth}{@{\extracolsep{\fill}}lccc}
    \toprule
    Criterion & \shortstack{GPT-5.4\\P / R (\%)}
      & \shortstack{Qwen3.5-4B FT\\P / R (\%)}
      & \shortstack{Qwen3.5-0.8B FT\\P / R (\%)} \\
    \midrule
    \multicolumn{4}{l}{\emph{Golden (selected checkpoints)}} \\
    Macro L1: concept & 74.08 / 75.34 & 84.21 / \textbf{91.43} & \textbf{85.60} / 89.82 \\
    Macro L2: + span & 52.44 / 51.65 & 72.62 / 72.93 & \textbf{75.47} / \textbf{75.70} \\
    Macro L3: + correctness & 45.77 / 45.08 & 64.48 / 64.82 & \textbf{67.84} / \textbf{68.07} \\
    \addlinespace
    \multicolumn{4}{l}{\emph{Golden2 (selected checkpoints)}} \\
    Macro L1: concept & 72.26 / 68.31 & 84.41 / 84.85 & \textbf{86.29} / \textbf{87.65} \\
    Macro L2: + span & 47.64 / 46.13 & 67.57 / 67.16 & \textbf{68.90} / \textbf{68.68} \\
    Macro L3: + correctness & 41.65 / 40.26 & 59.91 / 59.37 & \textbf{62.06} / \textbf{61.79} \\
    \bottomrule
  \end{tabular*}
  \caption{Key-macro precision and recall (\%) assigning equal weight to each distinct target
    matching key at each level, corresponding to the micro results in Table~\ref{tab:annotation-results}.
    Bold marks the best value for each metric in each row.}
  \label{tab:annotation-macro}
\end{table}

\subsection{Pre-Fixed Rare-Concept Subset}
\label{sec:rare-nine}
We selected the nine concepts with the fewest annotations in the raw 83,815-group
training split after restricting labels to the 34-concept inventory. Selection used no
benchmark outcomes; there was no tie at the cutoff. Table~\ref{tab:rare-nine} reports the
selection counts and gold support on the primary Golden2 evaluation benchmark (Golden serves
as an additional robustness check).

\begin{table}[!htbp]
  \centering
  \scriptsize
  \setlength{\tabcolsep}{3pt}
  \begin{tabular*}{\columnwidth}{@{\extracolsep{\fill}}lrr}
    \toprule
    Concept & Raw train & Golden2 \\
    \midrule
    Present perfect continuous & 187 & 5 \\
    Causative / permissive & 259 & 12 \\
    Past perfect & 334 & 9 \\
    Zero conditional & 335 & 13 \\
    Second conditional & 490 & 17 \\
    Reported speech & 574 & 19 \\
    Possessive \textit{'s} / whose & 825 & 46 \\
    One / ones & 1,008 & 20 \\
    Past continuous & 1,151 & 30 \\
    \midrule
    Total & 5,163 & 171 \\
    \bottomrule
  \end{tabular*}
  \caption{Pre-fixed Rare-9 concepts and annotation support. Raw-train counts determined
    subset membership; Golden2 counts are reported only as evaluation support.}
  \label{tab:rare-nine}
\end{table}

\subsection{Bootstrap Resampling and Statistical Significance}
\label{sec:stat-details}
Bootstrap resampling operates on packed five-utterance inference batches, so surplus
generated utterances remain assigned to the batch that produced them and are scored as
false positives. For paired ablation intervals, metric differences are computed within
each aligned cluster resample, taking the 2.5 and 97.5 percentiles across 20,000 draws.
For fixed-set Mean-34 and Rare-9 intervals, bootstrap draws without gold support for every
included concept were redrawn conditionally (Golden2: 138 draws for 4B and 153 for
0.8B; Golden: 769 draws for each size).

For pairwise comparisons against GPT-5.4, two-sided paired randomization tests permute
model predictions across systems at the five-utterance batch cluster level using
$B = 20{,}000$ Monte Carlo permutations. Empirical $p$-values are computed as
$p = (b + 1) / (B + 1)$, where $b$ is the number of permuted absolute test statistic
differences exceeding or equaling the observed difference; tests yielding zero exceedances
($b = 0$) are reported as $p < 0.0001$ ($p = 1/20{,}001$).

\FloatBarrier
\section{Serving Capacity, Autoscaling, and Cost}
\label{sec:serving-details}

\begin{figure}[!htbp]
  \centering
  \includegraphics[width=\columnwidth]{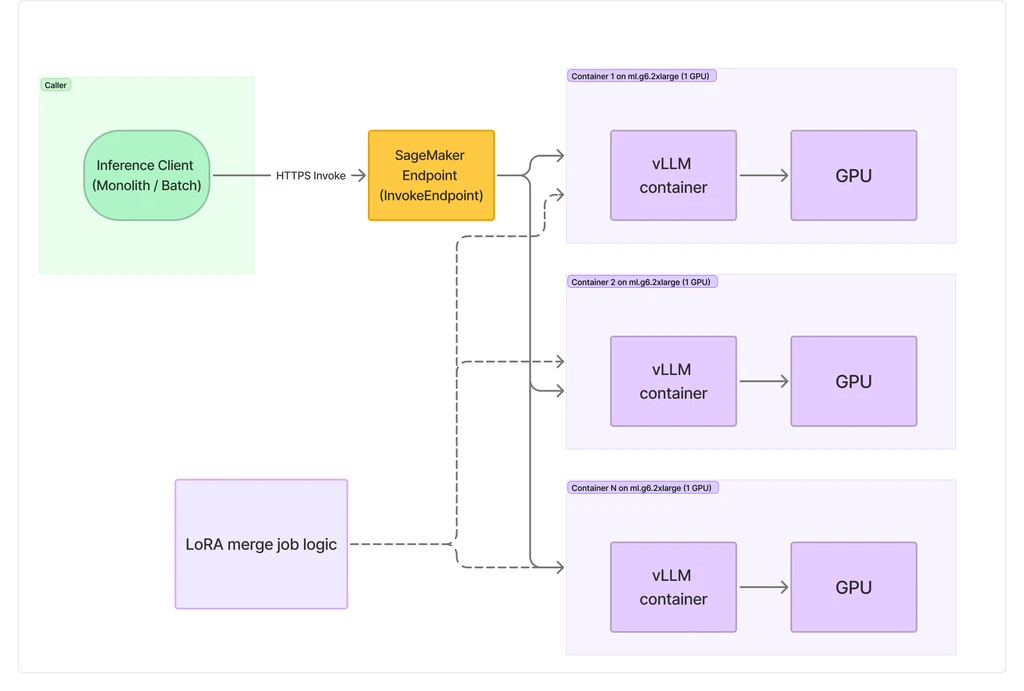}
  \caption{Simplified production \texttt{g6.2xlarge} native-vLLM topology. The application backend and
    internal inference API are collapsed into the inference-client block. SageMaker
    distributes calls across independently provisioned one-GPU containers.}
  \label{fig:serving-setup}
\end{figure}

Figure~\ref{fig:serving-setup} provides the detailed topology omitted from the
page-limited main paper. SageMaker distributes invocations across replicas; within each
replica, vLLM batches concurrent sequences using chunked prefill, prefix caching, and an
FP8 KV cache.

\paragraph{Workload and measured capacity.}
Before inference, the application partitions prepared utterance--correction pairs into
five-pair requests (averaging 4.27 pairs in production). To determine serving capacity,
concurrent clients replayed five-pair requests through a warmed inference service.
Eight \texttt{g6.2xlarge} instances sustained throughput above the observed post-rollout peak.

\subsection{Autoscaling Policy and Load Test}
\label{sec:autoscaling-validation}

The production endpoint uses AWS Application Auto Scaling targeting four concurrent in-flight
requests per instance, with 10-minute scale-out and 20-minute scale-in cooldowns across a range
of one to eight instances. We validated this policy by deploying the model to a staging endpoint
and replaying a staged load test that rose to a peak and then declined
(Table~\ref{tab:autoscaling-validation}).

\begin{table}[!htbp]
  \centering
  \scriptsize
  \setlength{\tabcolsep}{2.2pt}
  \begin{tabular*}{\columnwidth}{@{\extracolsep{\fill}}lrrl}
    \toprule
    Phase & p50 (s) & p99 (s) & Instances \\
    \midrule
    p1 & 4.43 & 6.87 & $0\rightarrow1$ \\
    p2 & 4.71 & 7.66 & $1\rightarrow2$ \\
    p3 & 4.53 & 20.07 & $2\rightarrow5$ \\
    p4 & 4.90 & 7.64 & $5\rightarrow5$ \\
    p5 & 4.23 & 6.46 & $5\rightarrow5$ \\
    p6 & 4.06 & 6.25 & $5\rightarrow4$ \\
    \bottomrule
  \end{tabular*}
  \caption{Predeployment autoscaling test. ``Instances'' gives the actual transition
    during each traffic phase; the initial zero denotes endpoint startup before the
    configured one-instance minimum became active. Latency is measured per request.}
  \label{tab:autoscaling-validation}
\end{table}

At the peak, five instances sustained traffic with a 7.64-second p99 latency, within
target concurrency thresholds. As load dropped, the scale-in cooldown retained capacity
before returning to the single-instance baseline.

\subsection{Cost Projection}
\label{sec:cost-model}

To project monthly serving expenditure under live traffic patterns, we analyzed 26 days of
hourly production request volume following full rollout. For each production hour, required
capacity was the observed request rate divided by a planning capacity set to 70\% of
measured saturation throughput (a 30\% headroom margin, consistent with the peak phase
in Table~\ref{tab:autoscaling-validation}, which held a steady instance count after latency
had settled), rounded up and clamped to the deployed 1--8
instance range.

Average provisioned capacity was multiplied by 730 monthly hours and public on-demand
SageMaker real-time pricing for \texttt{g6.2xlarge} instance-hours
\citep{aws-sagemaker-pricing-2026}. Compared against measured May 2026 production costs for
the low-reasoning GPT-5.4 configuration, the resulting SLM serving cost is 6.2\% of baseline
per lesson---an approximately 16$\times$ reduction.

This projection isolates recurring detection serving: the GPT-5.4 baseline is measured in
production, while the SLM estimate combines observed demand, load-tested throughput, and
public cloud pricing. Common upstream filtering and correction, fallback API calls, and
one-time training and deployment work are excluded. Absolute spend, per-lesson costs, and
eligible-lesson volumes are withheld.

\FloatBarrier
\section{Task Specifications and Auxiliary Probes}
\label{sec:task-specifications}

\subsection{Task Schema and Logical Contract}
\label{sec:task-schema-appendix}

Table~\ref{tab:task-schema} states the grammar concept annotation contract of
Section~\ref{sec:task} field by field.

\begin{table}[!htbp]
  \centering
  \small
  \setlength{\tabcolsep}{3pt}
  \begin{tabular*}{\columnwidth}{@{\extracolsep{\fill}}p{0.20\columnwidth}
      p{0.16\columnwidth}p{0.54\columnwidth}}
    \toprule
    Field & Type & Definition \\
    \midrule
    Original & Text & Learner utterance extracted from the ASR lesson transcript. \\
    Corrected & Text & Upstream LLM correction preserving the intended meaning. \\
    Concept & Enum & One of 34 target grammar concepts. \\
    Span & Text & Exact contiguous substring of the original utterance. \\
    Correct & Boolean & Whether the concept is used correctly in that span. \\
    Explanation & Text & Corrective feedback when usage is incorrect; empty otherwise. \\
    \bottomrule
  \end{tabular*}
  \caption{Logical input and output schema for grammar concept annotation. The first two
    fields form the input; the remaining fields describe each output annotation.}
  \label{tab:task-schema}
\end{table}

\subsection{Explanation Judgment Rubric}
\label{sec:llmj-rubric}

The LLM judge assigns \emph{accept} to correct and actionable feedback and \emph{pass} to
useful but pedagogically weak feedback. It assigns \emph{reject} when feedback targets the
wrong locus, states an incorrect rule, proposes an invalid fix, is tautological, or is
empty. Explanation acceptability (Section~\ref{sec:evaluation}) is defined as the proportion
of evaluated explanations receiving either \emph{accept} or \emph{pass}, grouping together
outputs that provide factually sound pedagogical assistance without misguiding the learner,
while isolating actively erroneous or misleading feedback under \emph{reject}.

In the human validation stage, an independent reviewer evaluated every model-generated
explanation under this rubric, and a second reviewer resolved discrepancies to establish
final aligned judgments. Across all three systems on Golden2, aligned human acceptability
rates converged to approximately 95\% with no statistically significant differences
($\chi^2$ test of independence, $p > 0.30$), in contrast to the uncalibrated LLM judge.

\subsection{Synthetic Distribution-Shift Probe}
\label{sec:synthetic-probe}
We tested the selected Qwen3.5-0.8B model on 200 learner-like utterances independently
authored and annotated by GPT-5.5. Concepts and correctness rates followed Golden2, while
topics spanned 15 domains; deterministic checks enforced schema, span, and label
constraints. Because the labels were teacher-generated, this is a diagnostic probe rather
than a third golden benchmark. \emph{Partial-span} matching requires at least 50\% character
overlap with the target span, whereas \emph{exact-span} requires exact string equality of the text span.

\begin{table}[!htbp]
  \centering
  \footnotesize
  \setlength{\tabcolsep}{4pt}
  \begin{tabular*}{\columnwidth}{@{\extracolsep{\fill}}lrr}
    \toprule
    Metric & Golden2 & Synthetic \\
    \midrule
    L1 concept micro F1 & 92.58 & 93.50 \\
    Partial-span F1 & 94.53 & 93.75 \\
    Exact-span F1 & 87.61 & 62.92 \\
    Parse errors & 1 / 2,977 & 0 / 200 \\
    \bottomrule
  \end{tabular*}
  \caption{Performance of the same Qwen3.5-0.8B checkpoint under real and synthetic
    evaluation distributions. Scores are percentages. Span axes evaluate localization
    independently of concept classification; exact-span performance is diagnostic because
    span-boundary conventions were not calibrated across annotators.}
  \label{tab:synthetic-shift}
\end{table}

As reported in Table~\ref{tab:synthetic-shift}, concept and partial-span F1
remained stable and all synthetic outputs parsed, but exact-span F1 fell by 24.69 points.
The synthetic-data generator used span-boundary rules that differed from those used
for the human-annotated benchmarks. The drop is therefore consistent with an
annotation-protocol mismatch.

\FloatBarrier
\end{document}